\pdfoutput=1
\documentclass[11pt]{article}
\usepackage{coling2020}
\usepackage{times}
\usepackage{url}
\usepackage{latexsym}
\usepackage{textcomp}
\usepackage{booktabs} 
\usepackage[flushleft]{threeparttable}
\usepackage{multirow}
\usepackage{graphicx}
\usepackage{enumitem}
\usepackage{array}
\usepackage{amssymb}
\usepackage{xspace}
\usepackage{mathtools}
\usepackage{xcolor}
\usepackage{pgfplots}
\pgfplotsset{compat=newest}
\usepackage{tabularx}
\usepackage{filecontents}
\usepackage{tikz}
\usepackage{subcaption}
\usepackage{color}
\usepackage{todonotes}
\usepackage{caption}

\colingfinalcopy 

\title{Best Practices for Data-Efficient Modeling in NLG:\\ How to Train Production-Ready Neural Models with Less Data}
\author{
  Ankit Arun\thanks{~~Author list alphabetical by last name.}, Soumya Batra, Vikas Bhardwaj, Ashwini Challa,  \\
 \textbf{Pinar Donmez, Peyman Heidari, Hakan Inan, Shashank Jain,} \\
  \vspace{0.15cm}
  \textbf{Anuj Kumar, Shawn Mei, Karthik Mohan, Michael White\thanks{~~Work done while on leave from Ohio State University.}}\\
  \vspace{0.15cm}
  Facebook \\
  {\tt \{ankitarun, sbatra, vikasb, ashwinichalla,}\\
  \tt{  pinared, peymanheidari, inan, shajain,}\\
  \tt{  anujk, smei, mkarthik, mwhite14850\}@fb.com}
  }

\begin{document}
\maketitle{}
\begin{abstract}
Natural language generation (NLG) is a critical component  in conversational systems, owing to its role of formulating a correct and natural text response. Traditionally, NLG components have been deployed using template-based solutions.
Although neural network solutions recently developed in the research community have been shown to provide several benefits, deployment of such model-based solutions has been challenging due to high latency, correctness issues, and high data needs. 
In this paper, we present approaches that have helped us deploy data-efficient neural solutions for NLG in conversational systems to production.  We describe a family of sampling and modeling techniques to attain production quality with light-weight neural network models using only a fraction of the data that would be necessary otherwise, and show a thorough comparison between each. Our results show that domain complexity dictates the appropriate approach to achieve high data efficiency. Finally, we distill the lessons from our experimental findings into a list of best practices for production-level NLG model development, and present them in a brief runbook. Importantly, the end products of all of the techniques are small sequence-to-sequence models (\texttildelow 2Mb) that we can reliably deploy in production.
\end{abstract}

\section{Introduction}
Task-oriented dialog systems are commonplace in automated systems that interact with end users, including digital assistants, technical support agents, and various website navigation helpers. An essential part in any task-oriented dialog system is \emph{natural language generation} (NLG), which consumes data, typically fed in the form of a \emph{dialog act}, and converts it into natural language output to be served to the end user. The natural language response of the NLG component should 1) contain all essential information, 2) be contextualized around the user request, and 3) be natural sounding. Such a system requires consideration for content planning, correctness, grammaticality, and naturalness.

NLG systems employed in commercial settings are typically based on template-based text generation techniques \cite{Reiter-Dale:2000,gatt2018survey,dale-commercial-sota:2020}. In these, humans author a minimal set of responses templates with placeholder slot values. These slots are later filled at runtime, with the dialog input. Although template-based NLG modules are appealing due to their deterministic nature,  inherent correctness, and low latency, they have major drawbacks: First, separate templates need to be authored for different response variations; this behavior is unfavorable for scaling. Second, templates authored for a particular domain are commonly not reusable. 
Lastly, no matter the complexity of the language instilled into templates, they form a strictly discrete set of responses, and therefore are bound to be limited in their response naturalness.

More recently, advances in neural-network-based (conditional) language generation prompted a new direction in NLG research \cite{novikova2017e2e,budzianowski2018multiwoz,chen2020few,Balakrishnan2019constrainednlg,peng2020few}.
The process is typically split into two steps: (1) serialization of input data into a flattened meaning representation (MR), and (2) using the neural generation model to generate a natural language response conditioned on the MR. The models are trained on data that includes $\langle$MR, response$\rangle$ pairs, and therefore they are able to not only generate desired responses for MRs in their training data, but they are also expected to form coherent responses for novel MRs, owing to the generalization ability of their machine learning (ML) backbone.

However, deploying neural NLG systems in an industry setting is quite challenging. First, it is not trivial to train a model that reliably presents its input data with the high fidelity required from a user-serving dialog system. Second, the models require much high-quality human-annotated data, which is resource intensive. Consequently, data annotation is a major limiting factor for scaling model-based NLG across domains and languages.

In this work, we detail our approach to production-level neural NLG, with a focus on scalability and data efficiency. Adopting the tree-structured MR framework introduced in Balakrishnan et al.~\shortcite{Balakrishnan2019constrainednlg}, which allows better control over generated responses, we train sequence-to-sequence RNN models ~\cite{sutskever2014sequence,bahdanau2014neural,cho2014learning} that can produce high-fidelity responses. We then employ a multitude of techniques for reducing the amount of  required data, primarily powered by eliminating the ``hidden'' redundancy by grouping data points with similar semantics into \emph{buckets}. We train models either on the reduced data, or after increasing the size of the dataset using a novel synthetic augmentation technique. We also employ large, pre-trained attention-based language models~\cite{lewis2019bart}, fine-tuning them on the same datasets, and then using novel methods to distill their knowledge into smaller sequence-to-sequence models. Further, we train models on data from multiple domains, showing gains over models trained on individual domains when the domains are semantically close together.
We conclude with a compiled list of best practices for production-level NLG model development based on our analyses, and we present it as a runbook.

\section{Related Work}

NLG from structured data has been an active research area for decades, facilitated of late by datasets like the E2E Challenge~\cite{novikova2017e2e}, MultiWoz~\cite{budzianowski2018multiwoz} and Conversational Weather~\cite{Balakrishnan2019constrainednlg}.
Recently, Seq2Seq models~\cite{Wen-etal-D15-1199,tgenbaseline,Balakrishnan2019constrainednlg,rao2019tree}, have become popular for their superior naturalness and simplicity. These models have achieved high performance on benchmarks like E2E challenge~\cite{novikova2017e2e} and WebNLG challenge~\cite{gardent-etal-2017-webnlg}. However, they require a lot of data making them resource-intensive to stand up and manage at scale. 

Our work introduces an approach for bootstrapping data-efficient NLG models by auto-annotating unlabelled examples using a large pretrained sequence de-noiser model known as BART ~\cite{lewis2019bart} fine-tuned on a small annotated dataset. Additionally, to increase data collection efficiency, we present several bucketing strategies, which enable a more uniform data collection process over the possible semantic space. We improve upon the BART auto-annotation technique by combining it with an innovative method of dynamic data-augmentation (DDA) and fine-tuning BART auto-annotation on a small subset of data sampled using a medium grained bucketing approach. We also carried out experiments to examine the effects of bucketing granularity combined with domain complexity.

In similar studies, pretrained GPT models ~\cite{radford2019language} were used by Chen et al.~\shortcite{chen2020few} and Peng et al.~\shortcite{peng2020few}, who fine-tune them on a small set of in-domain data, but they did not distill these models into ones suitable for production.

Interestingly, Wen et al.~\shortcite{wen2016multidomain} demonstrated that the structure of arguments in existing dialogues can be used to guide data collection for low-resource domain adaptation, which is similar to the bucketing strategies explored here.  Additionally, Shah et al.~\shortcite{shah-etal-2018-bootstrapping} introduce a dialogue self-play method where templates are instantiated with database values to create synthetic utterances, similar to our dynamic data-augmentation method; however, their instantiated templates are then rewritten by crowd-workers, whereas in our DDA method, crowd-sourced utterances are delexicalized and then re-instantiated with random values. Kedzie \& McKeown \shortcite{kedzie-mckeown-2019-good} also make use of a similar technique in their work on self-training for neural NLG; by comparison, we experiment with DDA in a wider variety of training scenarios.

\section{Experimental Approach}
\subsection{Data}
The experiments were conducted using 4 task-oriented datasets: a Conversational Weather dataset introduced in Balakrishnan et al.~\shortcite{Balakrishnan2019constrainednlg} and three additional datasets for the Reminder, Time, and Alarm domains. These four datasets were selected due to their varying level of complexity, which will be explained further in the results section.\footnote{The datasets can be found at {\color{blue}https://github.com/facebookresearch/DataEfficientNLG} } In addition, these domains provide a good representation of various arguments such as \texttt{tense}, \texttt{date\_time}, and \texttt{date\_time\_range} as well as range queries that are typically seen across conversational systems. Descriptive statistics of the datasets are shown in Table~\ref{tab:datasetstats}.

\begin{table*}[t!]
  \centering
  \small
    \begin{tabularx}{0.87\linewidth}{ c|c|c|c|c|c|c|c}
         \cline{1-8}
      \textbf{Domain}& \textbf{\# of Training} & \textbf{\# of CB}  &\textbf{\# of MB}   &\textbf{\# of FBQ}  &\textbf{\# of FB}  & \textbf{\# of Validation} &\textbf{\# of Test}\\
     \cline{1-8}
     \textbf{Weather} & 25390 & 2240  & 6406  & 20343 & 15456 & 3078 & 3121\\
     \cline{1-8}
     \textbf{Reminder} & 9716 & 68  & 562  &1907 &739 & 2794 & 1397\\
     \cline{1-8}
     \textbf{Time} & 5530 & 18  & 288  & 863 & 330 & 1529 & 790\\
     \cline{1-8}
     \textbf{Alarm} & 7163 & 26  & 126 & 286 & 188 & 2024 & 1024\\
     \cline{1-8}
    \end{tabularx}
  \vspace{-0.2cm}
   \caption{Number of examples in training, validation, and test sets for all domains in addition to number of different buckets in the training set. CB, MB, FBQ, and FB stand for coarse-grained, medium-grained, fine-grained combined with query, and fine-grained buckets, respectively. }
     \label{tab:datasetstats}%
     \vspace{-0.5cm}
\end{table*}%

All of the datasets use a tree structure to store the  meaning representation (MR) that has been discussed in Balakrishnan et al.~\shortcite{Balakrishnan2019constrainednlg}. If necessary, they use discourse relations (CONTRAST and JUSTIFY), which encompass a possible list of dialog acts (REQUEST, INFORM, etc.). The dialog acts contain a list of slot key-value pairs to be mentioned. The tree structures are used to present semantic information to the models after flattening. Examples of flattened MRs are shown in Table~\ref{tab:weather_mr-example} and Table~\ref{tab:mr-example}. The synthetic user queries and scenarios were generated by engineers, the annotated responses were created by human annotators following guidelines written by computational linguists. The responses were verified to be grammatical and correct by the linguists to ensure data quality.

\begin{table*}[bh]
\begin{center}
\small
\resizebox{1.0\textwidth}{!}{%
\begin{tabularx}{\linewidth}{l p{14.7cm}}
\hline 
\textbf{Query} &  How is the weather over the next weekend? \\
\hline
\textbf{Reference} &  Next weekend expect a low of 20 and a high of 45. It will be sunny on Saturday but it'll rain on Sunday. \\
\hline
 & \texttt{\color{blue}{\textbf{INFORM\_1[}}\color{red}{temp\_low[}\color{black}20\color{red}{]}\color{red}{ temp\_high[}\color{black}45\color{red}{]} \color{red}{date\_time[}\color{orange} colloquial[ \color{black}next weekend \color{orange}]\color{red}{]}\color{blue}\textbf{]} }\\
 \textbf{ Our MR} & \texttt{\color{black}{\textbf{CONTRAST\_1[}}}\\
 \textbf{(Tree-based}& \hspace*{0.5cm} \texttt{\color{blue}{\textbf{INFORM\_2[}}\color{red}{condition[}  \color{black}sun \color{red}{]} \color{red} {date\_time[}\color{orange} weekday[ \color{black}Saturday \color{orange}]\color{red}{]}\color{blue}\textbf{]}} \\
\textbf{Scenario)} &\hspace*{0.5cm} \texttt{\color{blue}{\textbf{INFORM\_3[}}\color{red}{condition[} \color{black}rain \color{red}{]} \color{red}{date\_time[}\color{orange} weekday[ \color{black}Sunday \color{orange}]\color{red}{]}\color{blue}\textbf{]} } \\
 &\texttt{\color{black}{\textbf{]}}}\\
\hline
 & \texttt{\color{blue}{\textbf{INFORM\_1[}} \color{red}{date\_time[}\color{orange} colloquial[\color{black}next weekend\color{orange}]\color{red}{]} \color{black} expect a low of  \color{red}{temp\_low[}\color{black}20\color{red}{]}\color{black}}\\
  &\hspace*{0.5cm} \texttt{ and a high of \color{red}{ temp\_high[}\color{black}45\color{red}{]} \color{black}.\color{blue}\textbf{]} }\\
 \textbf{Annotated} & \texttt{\color{black}{\textbf{CONTRAST\_1[}}}\\
 \textbf{Reference}& \hspace*{0.5cm} \texttt{\color{blue}{\textbf{INFORM\_2[}} \color{black} it will be \color{red}{condition[}\color{black}sunny\color{red}{]} \color{red} {date\_time[} \color{black}on \color{orange} weekday[\color{black}Saturday\color{orange}]\color{red}{]}\color{blue}\textbf{]}} \\
  & \hspace*{0.5cm} \texttt{\color{black}{but}}\\
 &\hspace*{0.5cm} \texttt{\color{blue}{\textbf{INFORM\_3[}} \color{black}it'll \color{red}{condition[}\color{black}rain\color{red}{]} \color{red}{date\_time[} \color{black}on \color{orange}   weekday[\color{black}Sunday\color{orange}]\color{red}{]}\color{blue}\textbf{]} } \\
 &\texttt{\color{black}{\textbf{.]}}}\\
\hline
\end{tabularx}}
\end{center}
\vspace{-0.4cm}
\caption{\label{tab:weather_mr-example} A training example with a discourse relation (bold black node). Blue nodes are the dialog acts, red nodes are the first level arguments under dialog acts and orange nodes are the second level arguments. Argument values at the leaf nodes and terminal tokens are in black.}
\vspace{-0.3cm}
\end{table*}

\subsection{Bucketing}
All our datasets present tree-structured input. We found the tree structure helpful in grouping the training examples in order to reduce biases in the model-generated responses because of imbalanced distribution and also to improve data efficiency. We investigated several bucketing strategies that assign scenarios into groups based on their tree structures and argument values at different levels of granularity. During data collection, we observed that compared to random, bucket-assisted gradual data collection improved model performance due to more exhaustive MR coverage.

\paragraph{Coarse-grained (CB)}
This bucketing strategy was the coarsest level of granularity. Under this strategy, the scenarios (MRs) are grouped using high-level argument names, which are at most two levels below the root node. For example, consider a second level argument such as \texttt{date\_time} that may have multiple nested arguments in different combinations.

In Coarse-grained bucketing, all variations deeper than \texttt{date\_time} were ignored. An example is shown in Table~\ref{tab:mr-example}, where in both the \texttt{INFORM\_1} and \texttt{INFORM\_2} dialog acts, despite different sub-arguments for the parent argument \texttt{date\_time}, variations are ignored. This strategy creates the smallest number of buckets. In spite of the high possible data efficiency using this method, models might exhibit worse performance due to limited MR coverage in the training data.

\paragraph{Medium-grained (MB)}
At this level of granularity, all sub-arguments were considered for creation of the bucket hashes. However, for certain pre-determined arguments/sub-arguments with small and finite variation, the argument name was replaced with its value. An example is the argument \texttt{tense}, which has 3 possible values; hence, when creating bucket hashes, we replace the \texttt{tense} argument with \texttt{tense\_past}, \texttt{tense\_present}, or \texttt{tense\_future}. This led to an increase in the bucketing space by the number of possible values for each such argument.

\begin{table*}[t!]
\begin{center}
\small
\resizebox{1.0\textwidth}{!}{%
\begin{tabularx}{\linewidth}{l p{14.5cm}}
\hline 
\textbf{Query} &  Do I have any reminder to buy milk? \\
\hline
\textbf{Reference} &  Yes, there are 3 reminders. The first two are, buy milk at 7 PM and tomorrow. There's 1 other reminder. \\
\hline
 & \texttt{\color{blue}{\textbf{INFORM\_1[}}\color{red} amount[ \color{black}3 \color{red}]\color{blue}\textbf{]}}\\
\textbf{ Our MR} &   \texttt{\color{blue}{\textbf{INFORM\_2[}}\color{red}{todo[}  \color{black}buy milk \color{red}{]} \color{red} {date\_time[}\color{orange} time[ \color{black}7 pm \color{orange}]\color{red}{]}\color{blue}\textbf{]}} \\
\textbf{(Scenario)} & \texttt{\color{blue}{\textbf{INFORM\_3[}}\color{red}{todo[} \color{black}buy milk \color{red}{]} \color{red}{date\_time[}\color{orange} colloquial[ \color{black}tomorrow \color{orange}]\color{red}{]}\color{blue}\textbf{]} } \\
 & \texttt{\color{blue}{\textbf{INFORM\_4[}}\color{red} amount\_remaining[\color{black} 1\color{red} ]\color{blue}\textbf{]}}\\
\hline
 & \texttt{\color{blue}{\textbf{INFORM[}}\color{black}Yes, there are \color{red}amount[ \color{black}3 \color{red}] \color{black}reminders .\color{blue}\textbf{]}}  \\
 \textbf{Annotated} & \texttt{\color{blue}{\textbf{INFORM[}}\color{black}The first two are, \color{red}{todo[} \color{black}buy milk \color{red}{]}\color{black} at  \color{red}{date\_time[}\color{orange}time[ \color{black}7 pm \color{orange}]\color{red}{]}\color{blue}\textbf{]}} \color{black} \\
\textbf{Reference} &and \texttt{\color{blue}{}{\textbf{INFORM[}}\color{red}{date\_time[}\color{orange} colloquial[ \color{black}tomorrow \color{orange}]\color{red}{]}\color{black}.\color{blue}\textbf{]}}\\
& \texttt{\color{blue}{}\textbf{INFORM[} \color{black}There's \color{red}amount\_remaining[ \color{black}1 \color{red}] \color{black}other reminder.\color{blue}\textbf{]}} \\
\hline
\textbf{Delexicalized} &  \multirow{2}{*}{Do I have any reminder to todo\_\_a ?} \\
\textbf{Query} &   \\
\hline
 & \texttt{\color{blue}{\textbf{INFORM\_1[}}\color{red} amount\color{blue}\textbf{]}}\\
\textbf{ Coarse\_grained} &   \texttt{\color{blue}{\textbf{INFORM\_2[}}\color{red}{todo} \color{red}{date\_time}\color{blue}\textbf{]}}\\
\textbf{Bucket Hash} & \texttt{\color{blue}{\textbf{INFORM\_3[}}\color{red}{todo} \color{red}{date\_time}\color{blue}\textbf{]} } \\
 & \texttt{\color{blue}{\textbf{INFORM\_4[}}\color{red} amount\_remaining\color{blue}\textbf{]} } \\
\hline
 & \texttt{\color{blue}{\textbf{INFORM\_1[}}\color{red}amount\color{blue}\textbf{]}}\\
\textbf{ Medium\_grained} &   \texttt{\color{blue}{\textbf{INFORM\_2[}}\color{red}{todo} \color{red}{date\_time[}\color{orange}time\color{red}{]}\color{blue}\textbf{]}} \\
\textbf{Bucket Hash} & \texttt{\color{blue}{\textbf{INFORM\_3[}}\color{red}{todo} \color{red}{date\_time[}\color{orange}colloquial[ \color{black}tomorrow \color{orange}]\color{red}{]}\color{blue}\textbf{]}} \\
 & \texttt{\color{blue}{\textbf{INFORM\_4[}}\color{red} amount\_remaining\color{blue}\textbf{]} } \\
\hline
 & \texttt{\color{blue}{\textbf{INFORM\_1[}}\color{red}amount[ \color{black}amount\_\_gr1 \color{red}]\color{blue}\textbf{]}}\\
\textbf{ Fine\_grained} &   \texttt{\color{blue}{\textbf{INFORM\_2[}}\color{red}{todo[} \color{black}todo\_\_a \color{red}{]} \color{red}{date\_time[}\color{orange} time[ \color{black}time\_\_a \color{orange}]\color{red}{]}\color{blue}\textbf{]}} \\
\textbf{Bucket Hash} & \texttt{\color{blue}{\textbf{INFORM\_3[}}\color{red}{todo[} \color{black}todo\_\_a \color{red}{]} \color{red}{date\_time[}\color{orange} colloquial[ \color{black}tomorrow \color{orange}]\color{red}{]}\color{blue}\textbf{]} } \\
 & \texttt{\color{blue}{\textbf{INFORM\_4[}}\color{red} amount\_remaining[ \color{black}amount\_remaining\_eq1 \color{red}]\color{blue}\textbf{]} } \\
\hline
\end{tabularx}}
\end{center}
\vspace{-0.4cm}
\caption{\label{tab:mr-example} A training example from the reminder domain with its corresponding coarse\_grained, medium\_grained and fine\_grained buckets. There are no discourse relations.}
\vspace{-0.3cm}
\end{table*}

In contrast to coarse-grained bucketing, the \texttt{INFORM\_2} and \texttt{INFORM\_3} dialog acts are grouped under different buckets as part of this strategy, since the \texttt{date\_time} parent argument has different sub-arguments. Moreover, for the \texttt{INFORM\_3} dialog act, the value of  sub-argument \texttt{colloquial} is retained. This implies that if there was another dialog act with the same shape as \texttt{INFORM\_3} dialog act but a different value for the \texttt{colloquial} sub-argument, it would have been grouped into a different bucket than \texttt{INFORM\_3}. This strategy increased the number of buckets compared to the CB case, improving coverage of different response variations. An example of medium-grained bucket hash appears in Table~\ref{tab:mr-example}.


\paragraph{Fine-grained (FB \& FBQ)}
In this strategy, the goal was to group cases into the largest possible number of buckets in which the surface form of the sentence was independent of the argument values decided by linguists (FB). There were three major differences compared with the medium-grained approach:  all argument values are considered, with partial delexicalization; argument values under the same argument name can be grouped; and uniqueness of argument values was tracked. 
For example, as shown in Table~\ref{tab:mr-example}, all argument values are considered, where the \texttt{todo} values are delexicalized, while \texttt{colloquial} is not. In addition, if the value of  \texttt{amount} or \texttt{amount\_remaining} is 1, then the surface form of the response might change, since a plural form should be used for numbers more than 1. Therefore, there are two groups of these argument values, one for values greater than 1 and one for the value of 1. Finally, the \texttt{todo} argument values in \texttt{INFORM\_1} and \texttt{INFORM\_2} are the same. Therefore, they are both delexicalized to \texttt{todo\_\_a}, which is to differentiate between cases where the \texttt{todo}s are different, since the model was allowed to aggregate based on arguments values to limit verbosity and increase naturalness. (If the values of \texttt{todo}s were different, they would have been delexicalized to \texttt{todo\_\_a} and \texttt{todo\_\_b}, resulting in a different bucket hash.) 

Our production models receive as input a combination of query and MR, in order to enable the possibility of conditioning the surface form of the response based on the query. Therefore, an additional level of bucketing can be achieved by delexicalizing the query (FBQ). For example, in Table~\ref{tab:mr-example}, the user has asked about a specific reminder, and the response confirms that by saying ``Yes'' at the beginning. (Saying ``Yes'' might have been unnecessary under a different query.) Since the queries in the datasets were generated synthetically, we could reliably delexicalize the query and consider the delexicalized query during bucket hash creation. 

\subsection{Metrics}
\label{sec_metrics}

We used various metrics to compare the performance of our proposed sampling and modeling approaches across experiments. 
Mainly, we focused on \textit{Tree Accuracy}, which is a binary metric indicating whether the tree structure in the response is correct \cite{Balakrishnan2019constrainednlg}. This metric checks whether the structural tokens in the response are the same as those in the input MR, modulo reordering at the sibling level (see Balakrishnan et al.'s paper for complete details).
Tree accuracy is also used in production to guard against hallucination: if tree accuracy fails, we fall back on templates to ensure correct response generation, even if it is less natural. In addition, we report \textit{BLEU Score} \cite{papineni2002} for all of the experiments.

Tree Accuracy is a binary metric and can change from 1 to 0 even if one structural token is missing, as intended. We noticed that tree accuracy can fluctuate considerably due to the random initialization of the layer weights and the randomization in mini-batch creation, even if trained on the same dataset with the same training parameters. (This might be due to the fact that the models are trained to optimize for token-level likelihood, not correctness.) To track the effectiveness of the proposed approaches in reducing these fluctuations and increasing \textit{Robustness}, we report the standard deviation of tree accuracy values based on 5 training instances for each experiment. The reported tree accuracy values for each experiment is the maximum one achieved in the same 5 runs.

Human evaluations were used as a qualitative method to raise red-flags in this study. For human evaluation, the authors rated the responses on Correctness and Grammaticality, defined as:

\begin{itemize}
    \item \textbf{Correctness:} Evaluates semantic correctness of a response. Authors check for hallucinations, missing attributes, attribute aggregation and sentence structure.
    \item \textbf{Grammaticality:} Checks for grammatical correctness of a sentence, which includes subject-verb agreement, word order, completeness, etc. 
\end{itemize}

We report an \textit{Acceptability} metric, which is the proportion of correct and grammatical responses sent for human evaluation. Due to annotator constraints, we devised a method to select the top 150 most differentiating examples from each domain's test set, in order to provide an understanding of the performance of each approach on the most challenging MRs. First, we categorized all of the test samples using the fine-grained (FB) bucketing technique. Then, for each bucket, the sample with the least number of correct (tree accuracy) responses across all of the experiments was selected if at least one approach responded correctly. Finally, the top 150 buckets with the least correct response were selected.

In our experience, if a model output fails the tree accuracy check it has always been wrong, but passing tree accuracy does not guarantee acceptability. 
Nonetheless, it should be noted that the reported acceptability numbers are significantly worse than with our production models, as they are focused on the most challenging MRs. 
The production weather models have had very high acceptability, so we do not report acceptability for the Weather domain due to some bandwidth constraints. 

To compare data-efficiency, we defined \textit{Data Reduction Rate} as the percentage of the initial training examples that can be saved (not used for training) using any of the presented approaches.

\subsection{Models}

The model architectures used in this study are either a sequence-to-sequence one with stacked LSTMs \cite{bahdanau2014neural} or derivatives of BART ~\cite{lewis2019bart} with stacked transformers. 

In the LSTM-based models, we use trainable 50$d$ GloVe~\cite{pennington2014glove} embeddings. Model weight are updated using an ADAM optimizer~\cite{kingma2014adam}.  For each experiment, we start with a learning rate of 0.01 and reduce it by a factor of 0.1 if validation loss does not decrease for 2 epochs. Our loss function is label smoothed CrossEntropy, where the beta parameter is between [0.01, 1]. Each model was trained for 100 epochs with a batch size of 32 and terminated when the validation loss stopped decreasing for 5 epochs.

For BART, we use the 6 layer BART-Base model in fp16 mode. This helps avoid the memory issues, which were faced with using the 12 layer BART model. For each experiment, we use ADAM as our optimizer with 300 warm-up steps. The starting learning rate of 3e-5 is reduced by a factor of 0.5  if validation loss does not decrease for 5 epochs.  Each model is trained for 100 epochs with a batch size of 4 and terminated when the validation loss stopped decreasing for 7 epochs. With all models we use a beam size of 1 to decrease latency.

\paragraph{LSTM-based Sequence-to-Sequence Model (S2S) }
Our main LSTM-based model has a single-layer encoder and a single-layer decoder. The dimensions of both encoder and decoder hidden states are set to 128 with 0.2 dropout. The input to the model is a concatenation of the user query and the meaning representation produced by the dialog management system.

\paragraph{Joint-training (JT)}
We experimented with a simple joint-training strategy for domains with similar responses, MRs, and semantics using the S2S architecture. The datasets were combined and a joint model was trained on them. 
Here, Alarm and Reminder are the two domains that are similar to each other and thus these were the domains we experimented with for joint-training.

\begin{table*}[b]
\begin{center}
\small

\begin{tabularx}{1.0\linewidth}{l|l|lll}
\cline{1-3}
\multirow{6}{*}{\textbf{Epoch 1}} & \textbf{Augmented} & \multirow{2}{*}{ Do I have any reminder to go shopping ?} \\
& \textbf{Query} &   \\\cline{2-3}
& & \texttt{\color{blue}{\textbf{INFORM\_1[}}\color{red} amount[ \color{black}8 \color{red}]\color{blue}{\textbf{]}} }\\
& \textbf{ Augmented} &   \texttt{\color{blue}{\textbf{INFORM\_2[}}\color{red}{todo[} \color{black}go shopping \color{red}{]} \color{red}{date\_time}[\color{orange}time[ \color{black}10 AM \color{orange}]\color{red}{]}\color{blue}{\textbf{]}} } \\
& \textbf{ MR } & \texttt{\color{blue}{\textbf{INFORM\_3[}}\color{red}{todo[} \color{black}go shopping \color{red}{]} \color{red}{date\_time[} \color{orange} colloquial[ \color{black}tomorrow \color{orange}]\color{red}{]}\color{blue}{\textbf{]}} } \\
&  & \texttt{\color{blue}{\textbf{INFORM\_4[}}\color{red} amount\_remaining[ \color{black}1 \color{red}]\color{blue}{\textbf{]}} } \\
\cline{1-3}
\multirow{6}{*}{\textbf{Epoch 2}} & \textbf{Augmented} & \multirow{2}{*}{ Do I have any reminder to run ?} \\
& \textbf{Query} &   \\\cline{2-3}
& & \texttt{\color{blue}{\textbf{INFORM\_1[}}\color{red} amount[ \color{black}4 \color{red}]\color{blue}{\textbf{]}} }\\
& \textbf{ Augmented} &   \texttt{\color{blue}{\textbf{INFORM\_2[}}\color{red}{todo[} \color{black}run \color{red}{]} \color{red}{date\_time}[\color{orange}time[ \color{black}6 PM \color{orange}]\color{red}{]}\color{blue}{\textbf{]}} } \\
& \textbf{ MR } & \texttt{\color{blue}{\textbf{INFORM\_3[}}\color{red}{todo[} \color{black}run \color{red}{]} \color{red}{date\_time[} \color{orange} colloquial[ \color{black}tomorrow \color{orange}]\color{red}{]}\color{blue}{\textbf{]}} } \\
&  & \texttt{\color{blue}{\textbf{INFORM\_4[}}\color{red} amount\_remaining[ \color{black}1 \color{red}]\color{blue}{\textbf{]}} } \\
\cline{1-3}
\end{tabularx}
\end{center}
\vspace{-0.4cm}
\caption{\label{tab:dda-example} Two examples of how a single training example is augmented randomly at each epoch.}
\vspace{-0.3cm}
\end{table*}

\paragraph{Dynamic Data Augmentation (DDA)}
To increase data efficiency, we carried out experiments using a limited number (1,3,5) of examples per bucket (coarse, medium, and fine) to determine at what training size the performance gains would plateau with more data collection. At very high data efficiency levels, we noticed that the model performance fluctuated significantly based on the argument values, which was unacceptable for a production system.

An initial idea was to pre-process the input and feed the delexicalized query and MR to the model. Although we could reliably delexicalize the user query during model training, it would have been very unstable to implement such a technique in production. In addition, there were concerns about added latency and higher complexity of the system. Therefore, we trained the model with the raw user query and with an MR in which argument values are lexicalized, which originally resulted in low data-efficiency in our production domains. 

We devised Dynamic Data Augmentation (DDA) as a new technique to provide robust model response with respect to changes in argument values using only a fraction of human-annotated responses. The idea is to randomly replace pre-processed tokens in the leaf nodes such as \texttt{todo\_a}, \texttt{time\_a}, etc.---as shown in the fine-grained example in Table~\ref{tab:mr-example}---with a value from a list of possible values, which are not expected to change the surface form of the sentence during mini-batch creation.

We used DDA to train on small datasets formed by sampling one or fewer examples per fine-grained bucket. In addition to higher data efficiency, such randomization should theoretically reduce the possibility of over-fitting. Similarly, DDA enables the 1PerBucket sampling technique in low-resource domains, resulting in a more uniform distribution of MR types during training. If the delexicalized query and MR shown in Table~\ref{tab:mr-example} are included in the training data, Table~\ref{tab:dda-example} demonstrates how DDA would augment the example differently at each epoch. 

\paragraph{BART Data Augmentation (BART+DDA)}
In BART auto-annotation, a small subset of data is sampled by selecting one example from each medium-grained bucket, followed by fine-tuning the BART model directly on this dataset. The fine-tuned BART model is then run on unlabelled scenario data, as part of the sequence-level knowledge distillation step described in the next section (S2S+KD), and the examples which match in tree structure with the input scenario are selected for training data augmentation. Sampling the small data using medium-grained bucketing introduces two issues. Firstly, although most response variations can be captured, the variations where words around argument slots change depending upon the argument value might be missed. Secondly, model performance is not robust to varying argument values. 

DDA solves both the above issues. In the BART+DDA approach, instead of directly fine-tuning the BART model on a small data, we fine-tune it on the dynamically augmented data. 

\paragraph{S2S+KD}
BART+DDA suffers from high latency and model size. In an effort to create production-quality models, we run the BART+DDA model on unlabelled scenario inputs, and select examples which match in tree structure. To auto-annotate unlabelled scenarios, we run a beam search of following beam sizes [1, 5, 10], and select the first response which passes the tree accuracy check. With larger beam sizes, even if the lower responses pass the tree accuracy, they often tend to be incorrect.

We then combine the synthetically generated examples labelled by BART with the golden human-labelled small data, and train a  S2S model on it. 
Using BART as a teacher model to train a smaller, faster S2S model is similar to Kim and Rush's \shortcite{kim2019seqkd} sequence-level knowledge distillation (KD) approach.

\paragraph{S2S+KD+DDA}
The S2S+KD model can make mistakes because the majority of its training data comes from synthetic data generated by the BART+DDA model. If the BART+DDA model makes a mistake on a particular scenario, these mistakes get amplified because it will be repeated when auto-annotating similar unlabelled scenarios. Even if golden human data has the correct response for these scenarios, it might not be enough to correct these mistakes. 
With S2S+KD+DDA, we solve this problem by fine-tuning the S2S+KD model using the DDA approach only on the gold human-labelled small data, as in recent self-training work for MT \cite{He2020Revisiting}.

\begin{figure}[t!]
\begin{center}
  \includegraphics[width=0.85\textwidth]{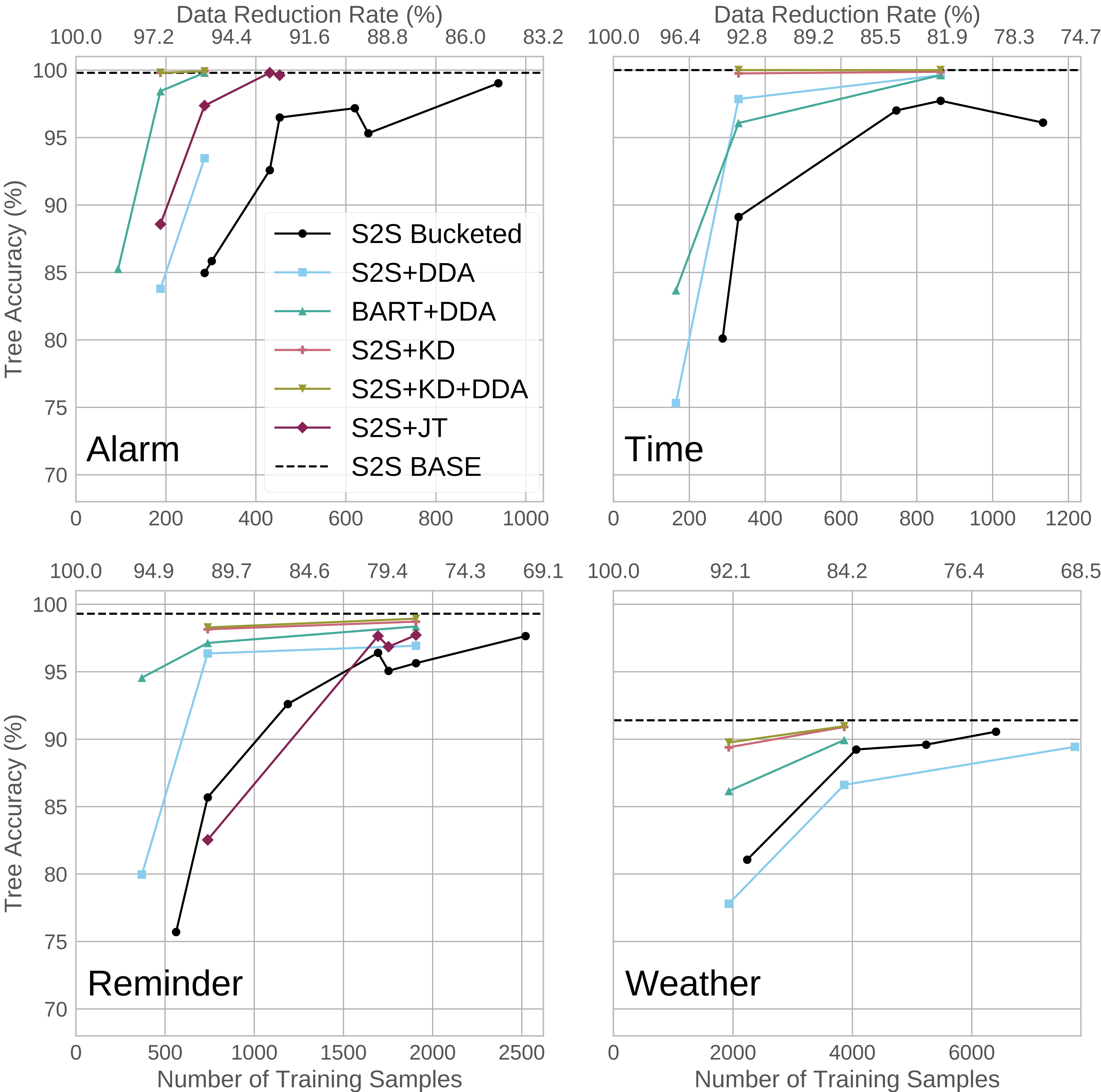}
  \vspace{-0.2cm}
  \caption{Change of tree accuracy vs.\ training data size and data reduction for the proposed approaches.}
  \label{fig:plot}
\end{center}
\vspace{-0.3cm}
\end{figure}

\section{Results and Discussion}
The results of selected experiments on Alarm, Time, Reminder and Weather domains are presented in Tables~\ref{tab:alarmresults}, \ref{tab:timeresults}, \ref{tab:reminderresults}, and~\ref{tab:weatherresults}, respectively. More comprehensive experimental results can be found in the Appendix. In addition, Figure~\ref{fig:plot} demonstrates comparative plots of all experiments with more than 70.0\% tree accuracy and a 70.0\% data reduction rate. From a pure data reduction point of view, the results suggest that S2S+KD+DDA and S2S+KD performed the best followed by BART+DDA. S2S+DDA generally improved performance compared to S2S trained on the same data. It can be also observed that joint domain training can improve performance compared to training only on the in-domain data.

The methods proposed and applied in the paper vary significantly in term of complexity and engineering effort needed to execute them. Therefore, we will analyze the results in each domain considering ease of use and scalability, with the major focus still on data reduction while maintaining performance.

BART+DDA, S2S+KD and S2S+KD+DDA improved the performance in all of the domains. However, deploying them to production was not justified for all the domains due to higher development and maintenance resources required. Specifically, BART+DDA has very high latency. S2S+KD and S2S+KD+DDA provide similar latency as the S2S variants. However, they require multiple engineering steps including training 3 models sequentially: First, fine-tune a BART+DDA model on a small dataset, followed by auto-annotating large amounts of unlabelled data for augmentation. Finally, S2S+KD and S2S+KD+DDA are trained sequentially on the augmented dataset. 

It is also not trivial to create non-annotated query and MR pairs for a new domain. Language expansion may also prove difficult as the cross-lingual mBART \cite{mBart} is only available for 25 languages, and cross-lingual extension of the auto-annotation techniques should be verified. KD variants are mostly beneficial where the required resources for annotating more data are high. Moreover, the gap between KD variants and S2S+DDA might diminish with the addition of 500-1000 examples. The balance between the required data annotation resources and the required engineering resources should be considered during approach selection based on the domain complexity.

\subsection{Alarm Domain}
Alarm was the least complex domain studied here. There were only 286 fine-grained (1PerFBQ) buckets, which were reduced to 188 buckets after the variation in user query was ignored (1PerFB). As shown in Table~\ref{tab:alarmresults} and Figure~\ref{fig:plot}, S2S Bucketed (1PerCB, 1PerMB, 1PerFB, and 1PerFBQ) experiments did not perform as well as other approaches. Interestingly, S2S+DDA did not perform as well as in other domains, which was probably due to the extremely low training data size in Alarm. However, combining Alarm with Reminder data improved the performance considerably (S2S+JT). S2S+JT tree accuracy reached within 0.2\% of the S2S BASE case with a data reduction rate of 93.7\%. The KD variants had the highest performance but they required higher development and maintenance resources.

\begin{table*}[t!]
  \small
   \centering
    \begin{tabularx}{1.0\linewidth}{c|c|ccccc}
    \cline{1-7}
     \multirow{2}{*}{\textbf{Data}}& \multirow{2}{*}{\textbf{Approach}} & \textbf{BLEU} & \textbf{\textsc{Tree}} & \multirow{2}{*}{\textbf{Acceptability}} & \textbf{\textsc{Data}}& \textbf{\textsc{TreeAcc}}\\ 
     &  & \textbf{\textsc{Score}} & \textbf{\textsc{Accuracy}} &  & \textbf{\textsc{Reduction}}& \textbf{\textsc{STDev}}\\
    \cline{1-7}
    \textbf{ALL}   & \textbf{\textsc{S2S BASE}} & 93.3 & 99.8 & - & 0.0 & 0.1 \\
     \cline{1-7}
    \textbf{5PerMB+Reminder} &  \textbf{\textsc{S2S+JT}}& 92.7 & \textbf{99.6} & 93.8 & \textbf{93.7} & 1.3 \\
     \cline{1-7}
     \textbf{1PerFBQ+Reminder} & \textbf{\textsc{S2S+JT}}& 92.6 & 97.6 & 83.5 & 96.1 & 2.8 \\
     \cline{1-7}
     \textbf{1PerFBQ} &  \textbf{\textsc{S2S}}& 91.6 & 85.0 & 67.0 & 96.1 & 16.2 \\
     \cline{1-7}
     \textbf{1PerFB} &  \textbf{\textsc{S2S}}& 92.3 & 91.8 & 75.0 & 97.4 & 28.6 \\
     \cline{1-7}
    \textbf{1PerFBQ} &  \textbf{\textsc{S2S+DDA}}& 92.7 & 93.4 & 77.3 & 96.1 & 14.3 \\
     \cline{1-7}
      \textbf{1PerFB} &  \textbf{\textsc{BART+DDA}} & 92.8 & 98.4 & 90.7 & 97.4 & 10.0 \\
    \cline{1-7}
    \textbf{1PerFB} &  \textbf{\textsc{S2S+KD}} & 93.2 & 99.8 & 92.8 &97.4 & 0.1 \\
    \cline{1-7}
    \textbf{1PerFB}  & \textbf{\textsc{S2S+KD+DDA}}& 93.2 & 99.8 & 93.8 & 97.4 & 0.1 \\
    \cline{1-7}
    \end{tabularx}
  \vspace{-0.2cm}
   \caption{Results on selected Alarm domain experiments in percentage. }
     \label{tab:alarmresults}%
     \vspace{-0.5cm}
\end{table*}%

\begin{table*}[b!]
  \small
  \centering
    \begin{tabularx}{0.85\linewidth}{ @{} c | c | c c c c c @{} }
    \cline{1-7}
     \multirow{2}{*}{\textbf{Data}}& \multirow{2}{*}{\textbf{Approach}} & \textbf{BLEU} & \textbf{\textsc{Tree}} & \multirow{2}{*}{\textbf{Acceptability}} & \textbf{\textsc{Data}}& \textbf{\textsc{TreeAcc}}\\ 
     &  & \textbf{\textsc{Score}} & \textbf{\textsc{Accuracy}} &  & \textbf{\textsc{Reduction}}& \textbf{\textsc{STDev}}\\
    \cline{1-7}
    \textbf{ALL}   & \textbf{\textsc{S2S BASE}} & 95.9 & 100 & - & 0.0 & 0.1 \\
     \cline{1-7}
     \textbf{1PerFBQ} & \textbf{\textsc{S2S}} & 95.4 & 97.7 & 90.0 & 85.0 & 2.5 \\
     \cline{1-7}
    \textbf{1PerFBQ}   & \textbf{\textsc{S2S+DDA}} & 95.7 & \textbf{99.6} & 96.7 & \textbf{85.0} & 4.8 \\
     \cline{1-7}
    \textbf{1PerFBQ}  & \textbf{\textsc{BART+DDA}} & 95.5 & 99.6 & 98.0 & 85.0 & 0.2 \\
    \cline{1-7}
    \textbf{1PerFBQ} & \textbf{\textsc{S2S+KD}} & 95.8 & 99.9 & 98.0 & 85.0 & 0.1 \\
    \cline{1-7}
    \textbf{1PerFBQ}  & \textbf{\textsc{S2S+KD+DDA}} & 95.8 & 100.0 & 99.3 & 85.0 & 0.1 \\
    \cline{1-7}
    \textbf{1PerFB}  & \textbf{\textsc{S2S+KD+DDA}}& 94.6 & 100.0 & 98.6 & 94.0 & 0.1 \\
    \cline{1-7}
    \end{tabularx}
  \vspace{-0.2cm}
   \caption{Results on selected Time domain experiments in percentage.}
     \label{tab:timeresults}%
\end{table*}%

\begin{table*}[b!]
  \small
  \centering
    \begin{tabularx}{0.92\linewidth}{ @{} c | c | c c c c c @{} }
    \cline{1-7}
     \multirow{2}{*}{\textbf{Data}}& \multirow{2}{*}{\textbf{Approach}} & \textbf{BLEU} & \textbf{\textsc{Tree}} & \multirow{2}{*}{\textbf{Acceptability}} & \textbf{\textsc{Data}}& \textbf{\textsc{TreeAcc}}\\ 
     &  & \textbf{\textsc{Score}} & \textbf{\textsc{Accuracy}} &  & \textbf{\textsc{Reduction}}& \textbf{\textsc{STDev}}\\
    \cline{1-7}
    \textbf{ALL} &  \textbf{\textsc{S2S BASE}} & 92.6 & 99.3 & - & 0.0 & 0.1 \\
    \cline{1-7}
     \textbf{5PerMB+Alarm}  & \textbf{\textsc{S2S+JT}} & 92.4 & 97.6 & 88.7 & 82.6 & 1.3 \\
     \cline{1-7}
     \textbf{1PerFBQ+Alarm}  & \textbf{\textsc{S2S+JT}} & 92.8 & 97.6 & 88.0 & 80.0 & 0.9 \\
     \cline{1-7}
     \textbf{1PerFBQ}   & \textbf{\textsc{S2S}} & 92.1 & 95.6 & 83.3 & 80.0 & 0.3 \\
     \cline{1-7}
     \textbf{1PerFB} & \textbf{\textsc{S2S}} & 90.4 & 85.7 & 47.3 & 92.5 & 27.0 \\
     \cline{1-7}
    \textbf{1PerFB} & \textbf{\textsc{S2S+DDA}} & 92.1 & \textbf{96.3} & 84.0 & \textbf{92.5} & 26.0 \\
    \cline{1-7} 
    \textbf{1PerFB}  & \textbf{\textsc{BART+DDA}} & 91.5 & 97.1 & 86.7 & 92.5 & 1.6 \\
    \cline{1-7}
    \textbf{1PerFB}  & \textbf{\textsc{S2S+KD}} & 91.9 & \textbf{98.1} & 89.3 & \textbf{92.5} & 0.2  \\
    \cline{1-7}
    \textbf{1PerFB}  & \textbf{\textsc{S2S+KD+DDA}} & 91.9 & \textbf{98.3} & 94.0 & \textbf{92.5} & 0.2  \\
    \cline{1-7}
    \end{tabularx}
  \vspace{-0.2cm}
   \caption{Results on selected Reminder domain experiments in percentage. }
     \label{tab:reminderresults}%
     \vspace{-0.5cm}
\end{table*}%

\subsection{Time Domain}
Time was the second simplest domain in this study. There were 860 fine-grained buckets, which were reduced to 330 buckets when the variation in user query was ignored. However, Time was unique in having tense as an argument which could result in errors that could pass tree accuracy.

S2S Bucketed (1PerCB, 1PerMB, 1PerFB, and 1PerFBQ) showed significantly lower performance both in terms of tree accuracy and acceptability compared to the S2S BASE experiment. S2S+DDA with 1PerFBQ data achieved tree accuracy of 99.6\%, which was just 0.4\% lower than S2S BASE (Figure~\ref{fig:plot}). In addition, S2S+DDA represented a data reduction of 85.0\%. Similar to Alarm, the KD variants perform the best but they require higher engineering resources.

\subsection{Reminder Domain}
Reminder was the second most complex domain with more than 1900 fine-grained buckets (1PerFBQ). As shown in Table~\ref{tab:reminderresults} and Figure~\ref{fig:plot}, S2S Bucketed  experiments with limited data (1PerCB, 1PerMB, 1PerFB, and 1PerFBQ) significantly under performed compared to the S2S BASE case. S2S+DDA with 1PerFB achieve tree accuracy of 96.3\% with data reduction of 92.5\%. However, a specific issue with change of ordering was detected during the human evaluations, which resulted in considerably low acceptability. A more comprehensive implementation of tree accuracy will be worked on to solved this issue. While joint-training increased both tree accuracy and acceptability over the S2S Bucketed experiments, other methods still outperformed joint-training.

S2S+KD and S2S+KD+DDA performed higher than other methods. Specifically, S2S+KD+DDA with a data reduction of 92.5\% achieved tree accuracy of 98.3\%, which was within 1.0\% of the S2S BASE experiment. Higher model maintenance resources will be required here as well, which might provide incentives for more data collection and ensuring higher data quality to improve the performance of S2S+DDA or S2S+JT to production levels.

\subsection{Weather Domain}
Weather was the most complex domain in this study. There were thousands of possible scenarios and the dataset size was considerably larger, accordingly. As demonstrated in Table~\ref{tab:weatherresults}, even a simple method such as S2S with only 1PerMB data achieved performance within 1.0\% of the S2S BASE case with a data reduction rate of 74.8\%. This is due to the high variety in the data that results in creation of more than 6400 medium-grained buckets, which was considerably higher than other domains. Therefore, we sub-sampled the buckets aggressively to examine the extent of possible data reduction.

Using only 1\//4 of the fine-grained buckets to train the models in the S2S+KD and S2S+KD+DDA approaches resulted in tree accuracy values within 2.0\% of the S2S BASE case. However, BART+DDA and S2S+DDA did not perform comparable to the KD variants. In addition, S2S+KD and S2S+KD+DDA provided low latency, which made the 92.5\% data reduction very favorable. In complex domains such as Weather, deploying models trained with more complex approaches that require higher development and maintenance resources is justified by high data-efficiency gains (19,000 fewer training samples here).

\begin{table*}[th!]
  \small
  \centering
    \begin{tabularx}{0.7\linewidth}{ @{} c | c | c c c c @{} }
    \cline{1-6}
     \multirow{2}{*}{\textbf{Data}}& \multirow{2}{*}{\textbf{Approach}} & \textbf{BLEU} & \textbf{\textsc{Tree}}  & \textbf{\textsc{Data}}& \textbf{\textsc{TreeAcc}}\\ 
      &  & \textbf{\textsc{Score}} & \textbf{\textsc{Accuracy}} & \textbf{\textsc{Reduction}}& \textbf{\textsc{STDev}}\\
     \cline{1-6}
     \textbf{ALL}   & \textbf{\textsc{S2S BASE}} & 91.4 & 91.4& 0.0 & 0.1 \\
     \cline{1-6}
     \textbf{1PerMB}  &\textbf{\textsc{S2S}} & 90.7 & 90.6 & 74.8 & 0.3 \\
     \cline{1-6}
    \textbf{0.5PerFB} & \textbf{\textsc{S2S+DDA}} & 89.8 & 86.6 & 85.0& 18.5 \\
     \cline{1-6}
      \textbf{0.25PerFB} & \textbf{\textsc{BART+DDA}}  & 89.2 & 86.2 & 92.5 & 1.8\\
     \cline{1-6}
      \textbf{0.25PerFB} & \textbf{\textsc{S2S+KD}}  & 89.7 & \textbf{89.4} & \textbf{92.5} & 0.1\\
     \cline{1-6}
      \textbf{0.25PerFB} & \textbf{\textsc{S2S+KD+DDA}}  & 89.8 & \textbf{89.8}  & \textbf{92.5} & 0.1\\
    \cline{1-6}
    \end{tabularx}
  \vspace{-0.2cm}
   \caption{Results on selected Weather domain experiments in percentage. }
     \label{tab:weatherresults}%
     \vspace{-0.5cm}
\end{table*}%

\section{Conclusions}
Several considerations are necessary for deploying model-based task-oriented dialogue systems to production. While increasing data efficiency was the primary goal of our study, we also considered and balanced data efficiency gains with several other factors such as acceptability, latency, and the required development and  maintenance resources. Focusing on four datasets for domains with varying level of complexity, we propose a sequential domain development run-book, where development of different domains can halt at different steps based on model performance evaluation. The steps are as follows:

\begin{itemize}
	\item Bucketing MRs based on a structure (tree-based here) in the data to avoid unnecessary and imbalanced data collection. Collect 1-3 examples per bucket. Train a model and evaluate it.
	\item If data for domains with similar tasks and semantics (like Reminder and Time) are available, Perform joint-training possibly followed by in-domain fine-tuning. Evaluate the model performance.
	\item Implement Dynamic Data Augmentation (DDA) to reduce the dependency of responses on interchangeable argument values. Train with DDA and evaluate the model performance.
	\item First, use pre-trained models (e.g. BART) to generate responses for unlabelled data. Then, combine the augmentation data with human-annotated data and train a small model (KD). Finally, fine-tune the model using DDA with the small human-annotated data. Evaluate the model performance.
	\item If necessary, collect more examples per MR bucket and start from the beginning to deploy the model with the lowest required development and  maintenance resources.
\end{itemize}

\section*{Acknowledgements}
We would like to thank our reviewers for their helpful feedback. Many thanks to our linguistic engineering team (Anoop Sinha, Shiun-Zu Kuo, Catharine Youngs, Kirk LaBuda, Steliana Ivanova, Ceci Pompeo, and Briana Nettie) for their hard work and for being great partners in this effort. We would also like to thank Jinfeng Rao, Kartikeya Upasani, Ben Gauthier, and Fiona Yee for their contributions.

\label{intro}

%
%
    %
    %
    %
    %
    %
    %
\newpage 
\bibliographystyle{coling}
\bibliography{coling2020}

\newpage
\section{Appendix}
\subsection{Detailed Experimental Results}

\begin{table*}[th!]
  \small
  \begin{center}
    \begin{tabularx}{0.99\linewidth}{c|c|ccccc}
    \cline{1-7}
     \multirow{2}{*}{\textbf{Data}}& \multirow{2}{*}{\textbf{Approach}} & \textbf{BLEU} & \textbf{\textsc{Tree}} & \multirow{2}{*}{\textbf{Acceptability}} & \textbf{\textsc{Data}}& \textbf{\textsc{TreeAcc}}\\ 
     &  & \textbf{\textsc{Score}} & \textbf{\textsc{Accuracy}} &  & \textbf{\textsc{Reduction}}& \textbf{\textsc{STDev}}\\
    \cline{1-7}
    \textbf{BASE}   & \textbf{\textsc{S2S}} & 93.3 & 99.8 & - & 0.0 & 0.1 \\
     \cline{1-7}
     \textbf{1PerCB} & \textbf{\textsc{S2S}} & 51.5 & 0.6 & 5.2 & 99.6 & 3.5 \\
     \cline{1-7}
    \textbf{1PerMB} & \textbf{\textsc{S2S}}& 87.0 & 56.2 & 38.1 & 98.3 & 23.9 \\
     \cline{1-7}
    \textbf{5PerMB+Reminder} &  \textbf{\textsc{S2S+JT}}& 92.7 & 99.6 & 93.8 & 93.7 & 1.3 \\
     \cline{1-7}
     \textbf{3PerFB+Reminder} & \textbf{\textsc{S2S+JT}}& 92.7 & 98.1 & 91.7 & 94.0 & 2.0 \\
     \cline{1-7}
     \textbf{1PerFBQ+Reminder} & \textbf{\textsc{S2S+JT}}& 92.6 & 97.6 & 83.5 & 96.1 & 2.8 \\
     \cline{1-7}
     \textbf{1PerFBQ} &  \textbf{\textsc{S2S}}& 91.6 & 85.0 & 67.0 & 96.1 & 16.2 \\
     \cline{1-7}
     \textbf{1PerFB} &  \textbf{\textsc{S2S}}& 92.3 & 91.8 & 75.0 & 97.4 & 28.6 \\
     \cline{1-7}
    \textbf{1PerFBQ} &  \textbf{\textsc{S2S+DDA}}& 92.7 & 93.4 & 77.3 & 96.1 & 14.3 \\
     \cline{1-7}
    \textbf{1PerFB}  & \textbf{\textsc{S2S+DDA}}& 91.6 & 83.8 & 67.0 & 97.4 & 20.9 \\
     \cline{1-7} 
     \textbf{1PerFBQ} &  \textbf{\textsc{BART+DDA}} & 92.5 & 99.8 & 93.8 & 96.1 & 1.3 \\
     \cline{1-7}
      \textbf{1PerFB} &  \textbf{\textsc{BART+DDA}} & 92.8 & 98.4 & 90.7 & 97.4 & 10.0 \\
    \cline{1-7}
    \textbf{1PerFBQ} &  \textbf{\textsc{S2S+KD}} & 92.9 & 99.9 & 94.8 & 96.1 & 0.1 \\
    \cline{1-7}
    \textbf{1PerFB} &  \textbf{\textsc{S2S+KD}} & 93.2 & 99.8 & 92.8 & 97.4 & 0.1 \\
    \cline{1-7}
    \textbf{1PerFBQ}  & \textbf{\textsc{S2S+KD+DDA}}& 92.9 & 99.9 & 94.8 & 96.1 & 0.05 \\
    \cline{1-7}
    \textbf{1PerFB}  & \textbf{\textsc{S2S+KD+DDA}}& 93.2 & 99.8 & 93.8 & 97.4 & 0.1 \\
    \cline{1-7}
    \end{tabularx}
  \end{center}
  \vspace{-0.2cm}
   \caption{Results on all Alarm domain experiments. All metrics are percentages. }
     \label{tab:alarmresultsall}%
     \vspace{-0.5cm}
\end{table*}%

 \vspace{0.5cm}
 
\begin{table*}[bh!]
  \small
  \centering
    \begin{tabularx}{0.85\linewidth}{ @{} c | c | c c c c c @{} }
    \cline{1-7}
     \multirow{2}{*}{\textbf{Data}}& \multirow{2}{*}{\textbf{Approach}} & \textbf{BLEU} & \textbf{\textsc{Tree}} & \multirow{2}{*}{\textbf{Acceptability}} & \textbf{\textsc{Data}}& \textbf{\textsc{TreeAcc}}\\ 
     &  & \textbf{\textsc{Score}} & \textbf{\textsc{Accuracy}} &  & \textbf{\textsc{Reduction}}& \textbf{\textsc{STDev}}\\
    \cline{1-7}
    \textbf{ALL}   & \textbf{\textsc{S2S BASE}} & 95.9 & 100 & - & 0.0 & 0.1 \\
     \cline{1-7}
     \textbf{1PerCB}   & \textbf{\textsc{S2S}} & 76.1 & 12.1 & 1.3 & 99.7 & 4.9 \\
     \cline{1-7}
     \textbf{1PerMB}  &\textbf{\textsc{S2S}} & 92.3 & 80.1 & 61.3 & 94.8 & 16.5 \\
     \cline{1-7}
     \textbf{1PerFBQ} & \textbf{\textsc{S2S}} & 95.4 & 97.7 & 90.0 & 85.0 & 2.5 \\
     \cline{1-7}
     \textbf{1PerFB}  & \textbf{\textsc{S2S}} & 93.5 & 89.1 & 76.7 & 94.0 & 7.6 \\
     \cline{1-7}
    \textbf{1PerFBQ}   & \textbf{\textsc{S2S+DDA}} & 95.7 & 99.6 & 96.7 & 85.0 & 4.8 \\
     \cline{1-7}
    \textbf{1PerFB}   & \textbf{\textsc{S2S+DDA}} & 94.9 & 97.8 & 87.3 & 94.0 & 10.2 \\
     \cline{1-7} 
     \textbf{1PerFBQ}  & \textbf{\textsc{BART+DDA}} & 95.5 & 99.6 & 98.0 & 85.0 & 0.2 \\
     \cline{1-7}
      \textbf{1PerFB}  & \textbf{\textsc{BART+DDA}} & 93.8 & 96.1 & 90.7 & 94.0 & 1.7 \\
    \cline{1-7}
    \textbf{1PerFBQ} & \textbf{\textsc{S2S+KD}} & 95.8 & 99.9 & 98.0 & 85.0 & 0.1 \\
    \cline{1-7}
    \textbf{1PerFB}  & \textbf{\textsc{S2S+KD}}  & 94.6 & 99.8 & 96.0 &  94.0 & 0.1\\
    \cline{1-7}
    \textbf{1PerFBQ}  & \textbf{\textsc{S2S+KD+DDA}} & 95.8 & 100.0 & 99.3 & 85.0 & 0.1 \\
    \cline{1-7}
    \textbf{1PerFB}  & \textbf{\textsc{S2S+KD+DDA}}& 94.6 & 100.0 & 98.6 & 94.0 & 0.1 \\
    \cline{1-7}
    \end{tabularx}
  \vspace{-0.2cm}
   \caption{Results on all Time domain experiments. All metrics are percentages. }
     \label{tab:timeresultsall}%
     \vspace{-0.5cm}
\end{table*}%

 \vspace{0.5cm}

\begin{table*}[bh!]
  \small
  \begin{center}
    \begin{tabularx}{0.99\linewidth}{c|c|ccccc}
    \cline{1-7}
     \multirow{2}{*}{\textbf{Data}}& \multirow{2}{*}{\textbf{Approach}} & \textbf{BLEU} & \textbf{\textsc{Tree}} & \multirow{2}{*}{\textbf{Acceptability}} & \textbf{\textsc{Data}}& \textbf{\textsc{TreeAcc}}\\ 
     &  & \textbf{\textsc{Score}} & \textbf{\textsc{Accuracy}} &  & \textbf{\textsc{Reduction}}& \textbf{\textsc{STDev}}\\
    \cline{1-7}
    \textbf{ALL} &  \textbf{\textsc{S2S BASE}} & 92.6 & 99.3 & - & 0.0 & 0.1 \\
     \cline{1-7}
     \textbf{1PerCB} &  \textbf{\textsc{S2S}} & 15.9 & 0.1 & 0.6 & 99.3 & 0.2 \\
     \cline{1-7}
    \textbf{1PerMB} & \textbf{\textsc{S2S}} & 89.94 & 75.7 & 28.0 & 94.2 & 22.4 \\
     \cline{1-7}
    \textbf{5PerMB+Alarm}  & \textbf{\textsc{S2S+JT}} & 92.4 & 97.6 & 88.7 & 82.6 & 1.3 \\
     \cline{1-7}
     \textbf{3PerFB+Alarm} & \textbf{\textsc{S2S+JT}} & 92.6 & 98.1 & 86.0 & 82.0 & 2.2 \\
     \cline{1-7}
     \textbf{1PerFBQ+Alarm}  & \textbf{\textsc{S2S+JT}} & 92.8 & 97.6 & 88.0 & 80.0 & 0.9 \\
     \cline{1-7}
     \textbf{1PerFBQ}   & \textbf{\textsc{S2S}} & 92.1 & 95.6 & 83.3 & 80.0 & 0.3 \\
     \cline{1-7}
     \textbf{1PerFB} & \textbf{\textsc{S2S}} & 90.4 & 85.7 & 47.3 & 92.5 & 27.0 \\
     \cline{1-7}
    \textbf{1PerFBQ}  & \textbf{\textsc{S2S+DDA}} & 92.3 & 96.9 & 82.0 & 80.0 & 0.2 \\
     \cline{1-7}
    \textbf{1PerFB} & \textbf{\textsc{S2S+DDA}} & 92.1 & 96.3 & 84.0 & 92.5 & 26.0 \\
     \cline{1-7} 
     \textbf{1PerFBQ} & \textbf{\textsc{BART+DDA}} & 92.1 & 98.3 & 93.3 & 80.0 & 0.2 \\
     \cline{1-7}
      \textbf{1PerFB}  & \textbf{\textsc{BART+DDA}} & 91.5 & 97.1 & 86.7 & 92.5 & 1.6 \\
    \cline{1-7}
    \textbf{1PerFBQ} & \textbf{\textsc{S2S+KD}} & 92.6 & 98.7 & 92.0 & 80.0 & 0.2  \\
    \cline{1-7}
    \textbf{1PerFB}  & \textbf{\textsc{S2S+KD}} & 91.9 & 98.1 & 83.3 & 92.5 & 0.2  \\
    \cline{1-7}
    \textbf{1PerFBQ}  & \textbf{\textsc{S2S+KD+DDA}} & 92.6 & 98.9 & 96.0 & 80.0 & 0.2  \\
    \cline{1-7}
    \textbf{1PerFB}  & \textbf{\textsc{S2S+KD+DDA}} & 91.9 & 98.3 & 94.0 & 92.5 & 0.2  \\
    \cline{1-7}
    \end{tabularx}
  \end{center}
  \vspace{-0.2cm}
   \caption{Results on all Reminder domain experiments. All metrics are percentages. }
     \label{tab:reminderresultsall}%
     \vspace{-0.5cm}
\end{table*}%

\begin{table*}[ht]
  \vspace{-47.5em}%
  \small
  \centering
    \begin{tabularx}{0.7\linewidth}{ @{} c | c | c c c c @{} }
    \cline{1-6}
     \multirow{2}{*}{\textbf{Data}}& \multirow{2}{*}{\textbf{Approach}} & \textbf{BLEU} & \textbf{\textsc{Tree}}  & \textbf{\textsc{Data}}& \textbf{\textsc{TreeAcc}}\\ 
      &  & \textbf{\textsc{Score}} & \textbf{\textsc{Accuracy}} & \textbf{\textsc{Reduction}}& \textbf{\textsc{STDev}}\\
     \cline{1-6}
     \textbf{ALL}   & \textbf{\textsc{S2S BASE}} & 91.4 & 91.4 & 0.0 & 0.1 \\
     \cline{1-6}
     \textbf{1PerCB}   & \textbf{\textsc{S2S}} & 88.1 & 77.9  & 91.2 & 2.6 \\
     \cline{1-6}
     \textbf{1PerMB}  &\textbf{\textsc{S2S}} & 90.7 & 90.6  & 74.8 & 0.3 \\
     \cline{1-6}
     \textbf{1PerFB}   & \textbf{\textsc{S2S}} & 91.3 & 91.4  & 40.0& 0.1 \\
     \cline{1-6}
    \textbf{1PerFB} & \textbf{\textsc{S2S+DDA}} & 91.3 & 91.1  & 40.0& 0.1 \\
     \cline{1-6}
    \textbf{0.5PerFB} & \textbf{\textsc{S2S+DDA}} & 89.8 & 86.6  & 85.0& 18.5 \\
     \cline{1-6}
    \textbf{0.25PerFB} & \textbf{\textsc{S2S+DDA}} & 87.3 & 77.8  & 92.5& 12.3 \\
     \cline{1-6} 
     \textbf{0.5PerFB} & \textbf{\textsc{BART+DDA}} & 90.2 & 89.9  & 85.0 & 1.7\\
     \cline{1-6}
      \textbf{0.25PerFB} & \textbf{\textsc{BART+DDA}}  & 89.2 & 86.2  & 92.5 & 1.8\\
    \cline{1-6}
    \textbf{0.5PerFB} & \textbf{\textsc{S2S+KD}} & 90.8 & 90.9 &  85.0 & 0.1\\
     \cline{1-6}
      \textbf{0.25PerFB} & \textbf{\textsc{S2S+KD}}  & 89.7 & 89.4  & 92.5 & 0.1\\
    \cline{1-6}
    \textbf{0.5PerFB} & \textbf{\textsc{S2S+KD+DDA}} & 90.8 & 91.0  & 85.0 & 0.1\\
     \cline{1-6}
      \textbf{0.25PerFB} & \textbf{\textsc{S2S+KD+DDA}}  & 89.8 & 89.8  & 92.5& 0.1\\
    \cline{1-6}
    \end{tabularx}
  \vspace{-0.2cm}
   \caption{Results on all Weather domain experiments. All metrics are percentages. }
     \label{tab:weatherresultsall}%
     \vspace{-0.5cm}
\end{table*}%

\end{document}